\documentclass[conference]{IEEEtran}

\ifCLASSINFOpdf
  \usepackage[pdftex]{graphicx}
  \graphicspath{{figures/}}
  \DeclareGraphicsExtensions{.pdf,.jpeg,.png}
\else
  \usepackage[dvips]{graphicx}
  \graphicspath{{../eps/}}
  \DeclareGraphicsExtensions{.eps}
\fi

\usepackage{amssymb}
\usepackage[cmex10]{amsmath}

\usepackage{algorithm}
\usepackage[noend]{algpseudocode}
\usepackage{hyperref}

\begin{document}
%
\title{Detecting Work Zones in SHRP 2 NDS Videos Using Deep Learning Based Computer Vision}
\author{

\IEEEauthorblockN{
Franklin Abodo, Robert Rittmuller, Brian Sumner and Andrew Berthaume
}\\
\IEEEauthorblockA{
Volpe National Transportation Systems Center\\
Office of Research, Development and Technology\\
U.S. Department of Transportation\\
Cambridge, MA 02142, USA\\
\{franklin.abodo, robert.rittmuller, brian.sumner, andrew.berthaume\}@dot.gov
}

}

\maketitle

\begin{abstract}
Naturalistic driving studies seek to perform the observations of human driver behavior in the variety of environmental conditions necessary to analyze, understand and predict that behavior using statistical and physical models. The second Strategic Highway Research Program (SHRP 2) funds a number of transportation safety-related projects including its primary effort, the Naturalistic Driving Study (NDS), and an effort supplementary to the NDS, the Roadway Information Database (RID). This work seeks to expand the range of answerable research questions that researchers might pose to the NDS and RID databases. Specifically, we present the SHRP 2 NDS Video Analytics (SNVA) software application, which extracts information from NDS-instrumented vehicles' forward-facing camera footage and efficiently integrates that information into the RID, tying the video content to geolocations and other trip attributes. Of particular interest to researchers and other stakeholders is the integration of work zone, traffic signal state and weather information. The version of SNVA introduced in this paper focuses on work zone detection, the highest priority. The ability to automate the discovery and cataloging of this information, and to do so quickly, is especially important given the two petabyte (2PB) size of the NDS video data set. 
\end{abstract}

\IEEEpeerreviewmaketitle

\section{Introduction}


Since the inception of SHRP 2, the presence or absence of work zones in a given NDS trip has been a desired piece of information. Prior efforts to conflate work zone occurrences with roadway information in the RID were not successful because they depended on 511 data provided by participating states. 511 data is only sparsely informative, indicating what segments of what highways had construction planned within a given time period. Whether actual construction equipment was present on the particular segment of highway over which a volunteer driver drove and at the time he or she drove is not an answerable question. In some cases, researchers have used 511 data to identify trips and their accompanying videos that supposedly contained a work zone, only to find themselves manually skimming through videos, sometimes finding what they were looking for and other times not \cite{smadi}. Further, one of the six participating states did not manage to supply 511-level data about work zones, making extraction of events from video the only option for those trips. This paper presents the SHRP 2 NDS Video Analytics (SNVA) software application, which aims to perform a complete and accurate accounting of work zone occurrences in the NDS video data set. To achieve this, SNVA  combines a video decoder based on FFmpeg, an image scene classifier based on TensorFlow, and algorithms for reading timestamps off of video frames, for identifying the start and end timestamps of detected work zone events, and for exporting those events as records in CSV files. We organize the presentation of SNVA as follows: Section  \ref{sec:deeplearningframeworkandarchitectureselection} discusses the motivations and methods behind our choice of deep learning framework and models; section \ref{sec:datasetconstructionandmodeldevelopment} details our approach to the joint development of the work zone detection model and the data set used to train it; and section \ref{sec:snvaapplicationdesignanddevelopment} describes our choice of hardware and software components, and highlights efforts made to optimize the video processing pipeline. In section \ref{futurework} we present the expected future directions of SNVA development, and we conclude the paper in section \ref{conclusion}.

\section{Deep Learning Framework and Architecture Selection} \label{sec:deeplearningframeworkandarchitectureselection}

\subsection{Deep Learning Framework Selection}

At the instantiation of the SNVA project, TensorFlow (TF) \cite{DBLP:journals/corr/AbadiABBCCCDDDG16} was identified as the deep learning framework most likely to contribute the most to the project's success.  We based this decision on two main factors. First, the apparent level of development and maintenance support as indicated by 1) the framework's popularity within machine learning research and practitioner communities as an open source tool, and 2) the framework's use in large-scale software applications by its creator Google. And second, the high-level API TensorFlow-Slim (TF-Slim), which was observed to include 1) useful demonstration code that could accelerate the team's learning and use of the framework, and 2) implementations of many CNN architectures accompanied by weights and biases pre-trained on ImageNet \cite{DBLP:journals/corr/RussakovskyDSKSMHKKBBF14} for use in transfer learning \cite{DBLP:journals/corr/YosinskiCBL14}.

\subsection{CNN Architecture Selection}

Convolutional neural networks (CNNs) have a widely demonstrated ability to apply to one task the weights and biases that were optimized for application to a different task. This is particularly true for tasks involving natural images such as those found in SHRP 2 NDS video data. Confident that an off-the-shelf CNN would prove suitable for the scene detection task, we set out to compare seven architectures for their in-sample test performance, inference speed and GPU utilization: InceptionV3 \cite{DBLP:journals/corr/SzegedyVISW15}, InceptionResnetV2 \cite{DBLP:journals/corr/SzegedyIV16}, MobilenetV1 \cite{DBLP:journals/corr/HowardZCKWWAA17}, MobilenetV2 \cite{DBLP:journals/corr/abs-1801-04381}, NASNet-Mobile \cite{DBLP:journals/corr/ZophVSL17}, ResnetV2-50 and ResnetV2-101 \cite{DBLP:journals/corr/HeZR016}. 

\begin{figure}[t!]  
\begin{center}  
\begin{tabular}{cc}  
\includegraphics[width=1.6in]{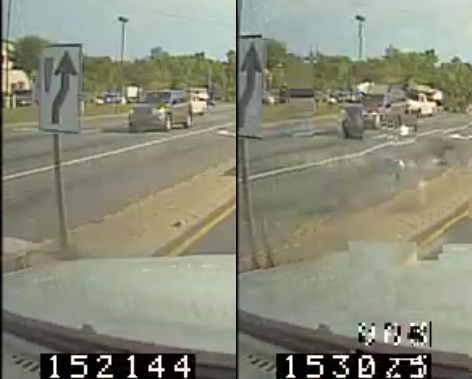}&
\includegraphics[width=1.6in]{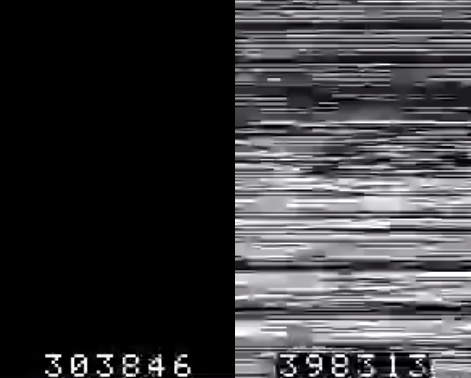}\\
(a) & (b)
\end{tabular}    
\begin{tabular}{cc}  
\includegraphics[width=1.6in]{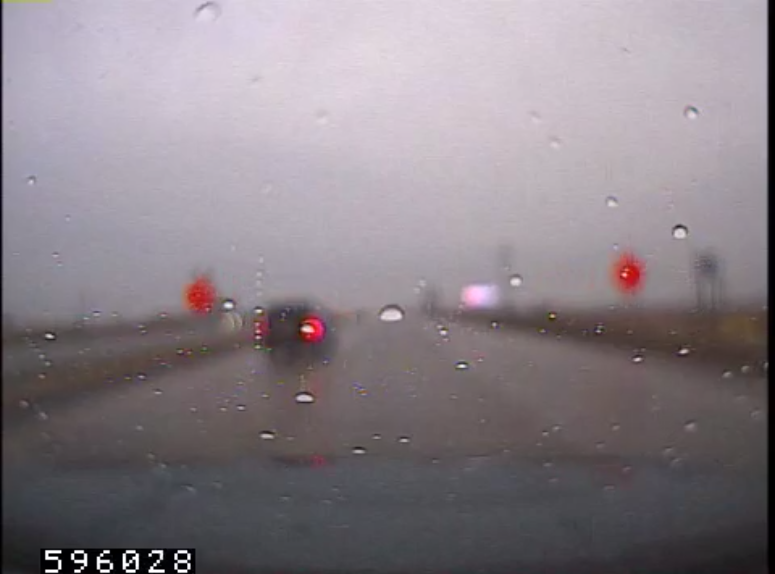}&
\includegraphics[width=1.6in]{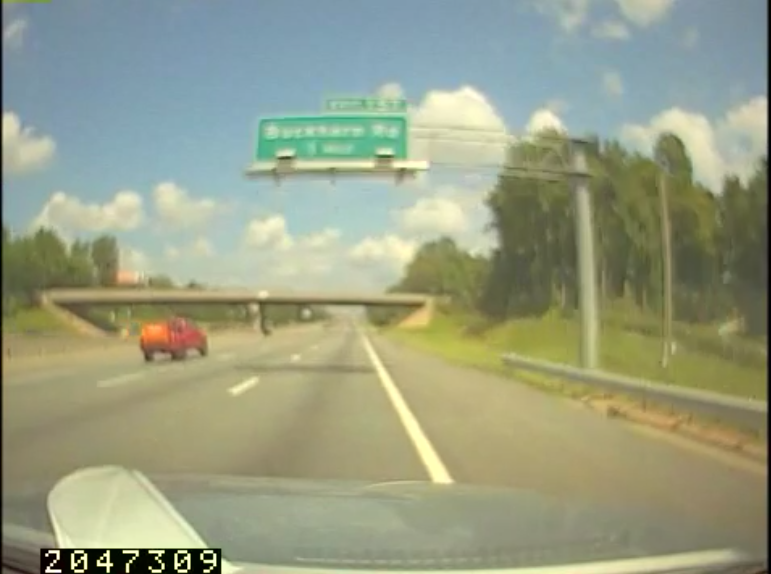}\\
(c) & (d)
\end{tabular} 
\caption{(a) The left frame is eligible for inclusion in the training set, but its successor is distorted and would thus be excluded from consideration. (b) Both the left and right frames' entire source videos would be excluded from training and validation sets. (c) Although the camera's focus is on the rain drops on the windshield, leaving the background blurry, this frame would be included in the training set with a label of \textit{warning sign} because the signs are sufficiently close to the vehicle to not be mistaken for background objects. The same signs at a slightly greater distance would easily not qualify. (d) This frame is also blurry because the camera is focused on a foreground object, but it would be included in the training set with a label of \textit{not work zone} because it is obvious that no work zone equipment is present in the scene.} 
\label{fig:questionablesamples}
\end{center}  
\end{figure}

\subsubsection{Validation Metric Selection}
In order to compare models learned using different architectures against one another during testing, and also against themselves during training for the purpose of early stopping, a validation metric is required. One simple and popular metric is accuracy: the ratio of correctly classified samples to the total number of samples. If we assume that class representation in the training and target data sets is imbalanced (e.g. the ratio of work zone scenes to non-work zone scenes is very low), then accuracy becomes an unreasonable metric. In a pathological example, if the ratio of work zone to non-work zone samples were 1:19, then a model could assign 100\% of test set samples to the not-work zone class and be 95\% accurate in spite of not detecting a single work zone.
With this in mind, we extended the TF-Slim demo code, which originally only measured accuracy, to add the following performance measures: precision, recall, \(F_1\), \(F_{0.5}\), \(F_2\), true and false positives and negatives, and total misclassifications. While all of the aforementioned metrics were used to develop an intuition about each model's performance, \(F_{0.5}\) was ultimately chosen as our single target measure. Recall that \(F\)-measures integrate precision and recall into a single metric, which is convenient when both measures are valuable. In the \(F_\beta\) formulation of the \(F\)-measure, setting \(\beta < 1\), \(\beta > 1\), or \(\beta = 1\) assigns more weight to precision, more weight to recall, or equal weight to precision and recall, respectively. In our use case, it is more important that the scene detector be correct when is claims to have discovered a work zone than to discover all existing work zones. We assume that the NDS data set is sufficiently large that some detections can be sacrificed for the benefit of relieving researchers of time wasted skimming through video clips that do not contain work zones. And so, we set \(\beta = 0.5\).

\subsubsection{Qualitative Filtering of Troublesome Samples}

The quality of SHRP 2 video data can vary on a number of dimensions. Camera or other hardware malfunctions can lead to discontinuities in video frames, completely noisy frames, completely black frames, frames in which the scene is rotated or frames that are out of focus, as illustrated in Figure \ref{fig:questionablesamples}. Work zone scene features of interest may be too distant to label confidently. Rain and other environmental impacts on vehicle windshields may distort features of interest, making them look identical to features normally observed in background scenes. In some cases we were required to exclude entire videos from consideration, but in most cases a few frames here and there were set aside. Only frames that could be labeled without hesitation were included in the data set, with the hope that the absence of excluded frames during training would lead the model to classify frames similar to them with low confidence at test time. If the assumption were to hold, a smoothing algorithm could be applied to perturb low confidence frames to match the class of their highly confident neighbors, potentially correcting misclassifications caused by the very exclusion of such frames during training. This exact behavior was observed in a handful of randomly sampled outputs during the testing of the SNVA application. 

\subsubsection{Transfer Learning using Weights and Biases Pre-trained on ImageNet}

Conscious of the seemingly infinitesimally small amount of training data available to us, we only considered CNN architectures for which weights and biases pre-trained on the ImageNet 2012 Large Scale Visual Recognition Challenge data set \cite{DBLP:journals/corr/RussakovskyDSKSMHKKBBF14} were available for download. The transferring of such weights and biases from one task to another has been demonstrated to aid in prediction tasks across a wide range of applications and scientific disciplines, particularly when the number of training samples is very low \cite{Ching20170387}. We compared training time and out-of-sample test performance for several CNNs initialized using random weights versus pre-trained weights and found that all showed better performance in both cases when weights were transferred. For the CNN competition and for further development of the selected architectures, the two-phase strategy for transfer learning presented by the TF-Slim authors was adopted \cite{website:tf-slim}, with the additional touch of keeping a few of the earliest layers frozen during fine-tuning as advised in \cite{DBLP:journals/corr/YosinskiCBL14}.

\subsubsection{CNN Competition and Results}
The objective of the CNN selection competition was to identify the single best candidate for inclusion in the final application. Architectures were compared using in-sample \(F_{0.5}\) scores, inference speed in frames per second, and GPU core and memory utilization. In our experiments, we identified MobilenetV2 as the most suitable candidate because of its combination of highest inference speed, lowest memory consumption, and relatively high \(F_{0.5}\) measure. The low memory consumption is of particular value because it permits either a large batch size or the concurrent assignment of multiple video processors to a single GPU. The competition results for all CNNs is presented in Table I.

\begin{table}[ht]
\begin{center}
\caption {CNN Architecture Performance Comparison} 
\centering
\begin{tabular}{|l|c|c|c|c|c|c|}
\hline
 & \textit{F0.5} & \textit{FPS} & \textit{GPU} & \textit{GPUMem} & \textit{BatchSz} & \textit{Steps} \\
\hline
\textit{IV3} & 0.971 & 783 & 96\% & 8031MB & 32 & 47.4K \\
\textit{IRV2} & 0.957 & 323 & 96\% & 7547MB & 64 & 41.9K \\ 
\textit{MV1} & 0.960 & 1607 & \textbf{91\%} & 8557MB & 32 & 45.7K \\
\textit{MV2} & 0.968 & \textbf{1615} & 94\% & \textbf{2413MB} & 32 & 45.5K \\
\textit{NM} & 0.964 & 1211 & 98\% & 2459MB & 128 & 45.8K \\
\textit{RV2-50} & \textbf{0.972} & 1000 & 98\% & 8543MB & 64 & 46.7K \\
\textit{RV2-101} & 0.931 & 645 & 98\% & 8543MB & 128 & 46.4K \\
\hline
\end{tabular}
\end{center}
\label{tab:cnn_architecture_performance_comparison}
\end{table}

\section{Data Set Construction and Model Development} \label{sec:datasetconstructionandmodeldevelopment}

In this section, we describe the methods used to jointly develop the selected model defined in the previous section together with the data set used to train, validate and test that model. Specifically, we discuss the processes by which we 1) determined what sources of data should contribute to data set construction, 2) defined a policy for excluding "unreasonable" data samples, 3) selected the CNN architectures that would compete for use in the final version of SNVA and the deep learning framework in which those architectures would be implemented, and 4) 
jointly developed the selected and training set.

\subsection{Data Source Selection}

Because the SHRP 2 NDS video data was collected using homogeneously instrumented vehicles, we expected the SNVA application to target videos that were consistent in quality and characteristics (e.g. resolution, camera focal length and other intrinsic properties, et cetera). In turn, we limited our sources for data set construction to the NDS videos themselves, assuming that images from publicly available data sets that happened to contain construction features would be too out-of-sample to be useful in classifying the target distribution of scenes.

A total of 1344 videos containing 31,535,862 frames were made available for use as sources for data set construction, including training, validation and test subsets. By manual inspection, the videos were observed to contain a variety of environmental scenes and features such as light and heavy rain, snow and fog, sunlight in front of, above and behind the subject vehicle, dusk, dawn and nighttime scenes, and highway and city scenes. This variety gave us confidence that our data source scene distribution was representative of the target scene distribution, in spite of constituting less than 0.0001 percent of the estimated total number of frames in the target data set. Examples of the variety of scenes are presented in Figure 2.

\begin{figure}[t!]  
\begin{center}  
\begin{tabular}{cc}  
\includegraphics[width=1.6in,height=1.2in]{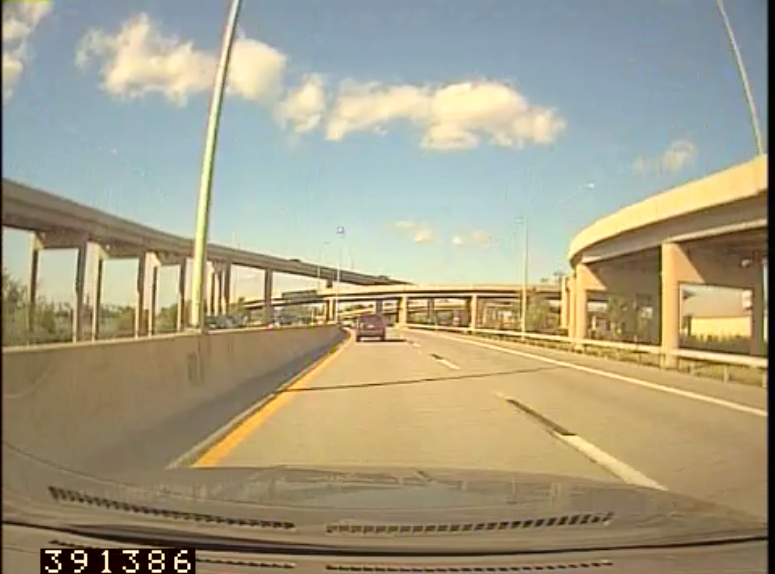}&
\includegraphics[width=1.6in,height=1.2in]{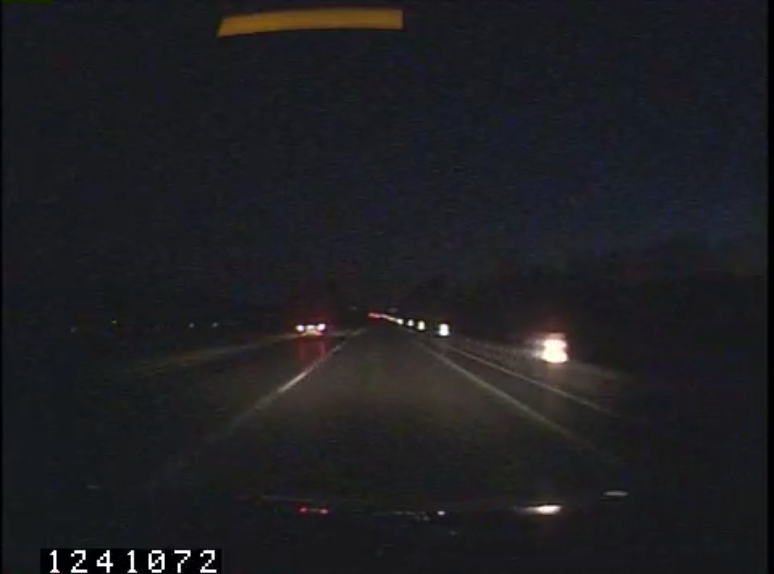}\\
(a) & (b)
\end{tabular}   
\begin{tabular}{cc}  
\includegraphics[width=1.6in,height=1.2in]{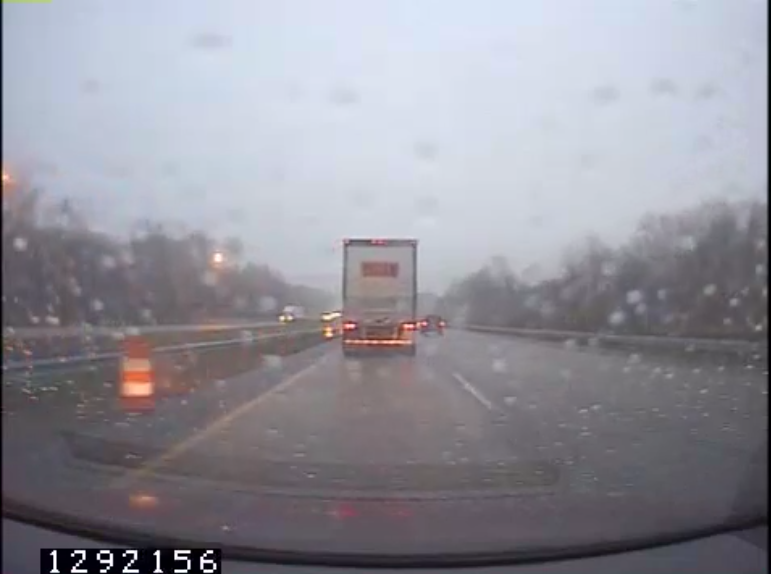}&
\includegraphics[width=1.6in,height=1.2in]{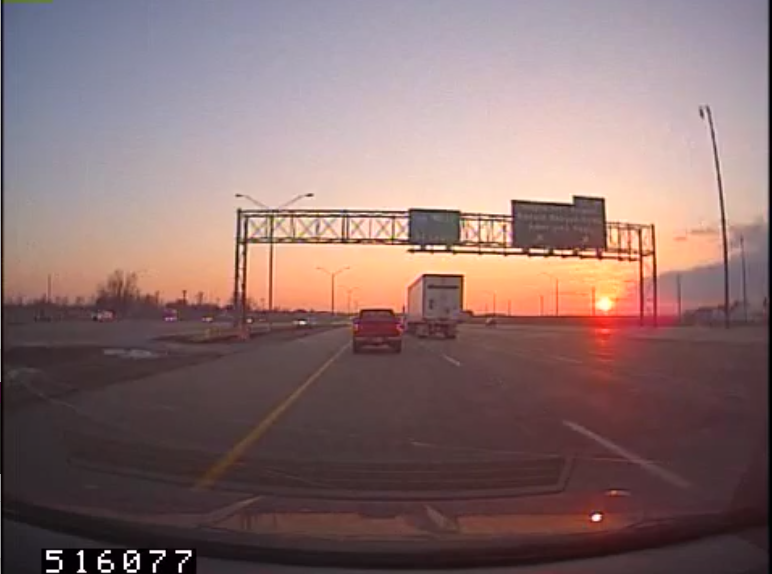}\\
(c) & (d)
\end{tabular}  
\caption{A variety of environmental conditions were present in the small subset of videos use for model development. (a) Clear-skied daytime with the sun behind the camera. (b) Nighttime with the subject vehicle's headlights illuminating construction drums. (c) Rainy daytime with the camera correctly focused on distant objects. (d) Dusk with the sun in front of the camera.}  
\end{center}  
\label{fig:environmentalconditionvariety}
\end{figure}

\subsection{Active Learning via Uncertainty Sampling}

Active learning is a set of semi-supervised techniques used to reduce the cost of machine learning model development. While there exist many varieties of active learning, all share the common objective of minimizing the number of training samples that require hand-labeling by a human, while still producing a model that meets inference performance requirements. For our purposes, we adopt a simple and commonly used method based on uncertainty sampling \cite{uncertainty-sampling}. In this approach, a model is initially trained on a small hand-labeled training set of "seed" samples, then used predict the classes of the remaining unlabeled samples. Samples for which the model is most uncertain in its label prediction (e.g. for which the probability distribution over classes is closest to uniform) are assumed to be the most informative to the model and are selected for inclusion in the next round of training, followed again by inference. This procedure is repeated until either 1) financial or human capital is exhausted, or 2) the model's performance converges (e.g. its lowest confidence prediction is above some desired threshold, or the number of uncertain samples per round  stops decreasing). One can think of these most uncertain samples as being most informative to the model because their inclusion in the next round of training would adjust the model's decision boundaries the most. Seen from another perspective, if a model is confident in its prediction of an unlabeled sample's class, then adding that example to the training set will not contribute to improving the model's performance (assuming the prediction is correct) because it would not likely adjust the model's decision boundaries.

Of course, when the model makes highly confident predictions about the wrong class, then it would be beneficial to include the affected samples in the next round of training. This point raises a dilemma; the only way to observe these misclassifications is to inspect all predictions and not just the uncertain ones, which negates the premise of active learning entirely. The resolution to this dilemma is another assumption; that if we focus on hand-labeling only the uncertain samples, eventually the model will learn enough from them that it will either 1) correct its highly confident misclassifications or 2) decrease its confidence in the wrong class enough so that a human ends up labeling the sample directly as part of the routine active learning process. This second behavior was observed during our labeling effort, but not rigorously studied. 

There remains a question of what probability threshold should mark the boundary between certain and uncertain class predictions. To begin to develop an answer, consider that in the most extreme case, no matter how many classes exist in a given classification problem, no more than one class can have a confidence greater than 0.5. Thus, one of the uncertainty boundaries should be 0.5. Anything below this is definitely in need of a human-assigned label. For the upper bound, above which the model is considered certain in its prediction, the closer that threshold is to 0.5 or 1.0, the lower or higher the number of examples proposed for hand-labeling will be, respectively. In the absence of any analytic method for determining the value of the upper threshold, we take a stratified approach and define five ranges (above 0.5) within which data points may be binned: (0.5, 0.6], (0.6, 0.7], (0.7, 0.8], (0.8,  0.9] and (0.9, 1.0]. Following this approach, the amount of effort devoted to hand-labeling can be determined dynamically as a function of the number of samples in each bin. At a minimum, every sample in the [0.0, 0.5] is automatically labeled. The decision to hand-label points in the next higher bin can be made one step at a time. In our application of this strategy, we started with a seed set of ~100,000 frames selected from eight videos. After the first round of training and inference over unlabeled samples, we selected an additional 350 from our total of 1314 videos to serve as sources for data set construction. Among these 350 videos, ~50,000 frames were binned in the range [0.0, 0.5] and we selected those for labeling. For the second round, we were able to expand the range to include (0.5, 0.6] as only ~30,000 were contained therein. We are at the time of this writing continuing to increase the training set size and improve the model while beta versions of the SNVA application are being tested in the deployment environment at VTTI.

Active learning is particularly attractive for our use case and critical to our success for two reasons. First, because work zone features are expected to occur in the target data set relatively infrequently, a simple uniform random sampling of video frames for inclusion in the training set would likely 1) ignore useful work zone-containing frames, and 2) result in time wasted labeling frames with redundant information content. Second, because the number of available unlabeled frames approaches 2\textsuperscript{25}, discovering all work zone-containing frames by exhaustively skimming through videos is not feasible.

\section{SNVA Application Design and Development} \label{sec:snvaapplicationdesignanddevelopment}

\subsection{Core Software Components}
\subsubsection{TensorFlow and the TF-Slim Image Classification Model Library}
The TensorFlow framework, together with the communities internal and external to Google that support it, are the primary enabler of this work. The public availability of quality demonstration code by way of the TF-Slim image classification library \cite{website:tf-slim} accelerated experimentation with and development of neural network models substantially. In addition to pre-trained models, the library included Python scripts for the training and evaluation of models, the creation of TFRecord-based datasets, and the conversion of large model checkpoint files into compact, constant operation-only protobuf files that are optimized for inference. We were able to easily extend the code to support project-specific needs such as:
\begin{enumerate}
	\item the incremental training data set construction method used in active learning,
    \item the augmentation of existing architecture implementations to support the NCHW format,
    \item the addition of command-line parameters to make the running of multiple training and evaluation scripts concurrently across multiple GPUs and CPUs convenient, and
    \item the addition of channel-wise standardization based on data set-level statistics as an optional preprocessing function.
\end{enumerate}
Another surprisingly useful component of the TF ecosystem was Tensorboard, a visualization tool that helped us monitor training and evaluation, compare the performance of various architectures during the competition mentioned in section \ref{sec:deeplearningframeworkandarchitectureselection}, and study and understand the structure and running state of TF-Slim models. 

\subsubsection{The FFmpeg Video/Audio Conversion Program}
The decoding of the MPEG-4 formatted videos into raw bytes for ingestion by analyzers was performed using FFmpeg. The program was also used to extract individual frames from videos and save them to disk for use in model development.
\subsubsection{The Numpy Scientific Computation Library}
The Numpy scientific computation library was exercised heavily in the algorithms that process video frame timestamps and that apply weighted averaging to smooth class probability distributions output by SNVA's models. The library's support for broadcasting and vectorization sped up operations noticeably when used in place of the naive/intuitive implementations that preceded them.
\subsubsection{The Python Programming Lanugage}
The SNVA application was implemented entirely in Python, making it easy to integrate the software components used. TF's model development API is written in Python, and the FFmpeg binary is easily invoked and interacted with using Python's subprocess module. Given Python's popularity in the data science community, we expect its use to help make this project accessible to would-be beneficiaries inside and outside of the authors' affiliated organizations.

\begin{algorithm}
\caption{ConvertTimestampImagesToStrings(\textit{T}, \textit{h}, \textit{w})}\label{alg:convtc}
\begin{algorithmic}[1]
\State $l \gets \textsc{Len}(T)$
\State $n\gets  w \div h$
\State $M \gets \textsc{GetTimestampDigitMaskArray}()$
\State $M \gets \textsc{Tile}(M, n)$
\State $M \gets \textsc{Transpose}(M, (0, 2, 1))$
\State $M \gets \textsc{Reshape}(M, (l, n, h, h))$
\State $T \gets \textsc{BinarizeTimestampImages}(T)$
\State $T \gets \textsc{Reshape}(T, (l, n, h, h))$
\State $T \gets \textsc{ExpandDims}(T, 1)$
\State $E \gets \textsc{Equal}(T, M)$
\State $A \gets \textsc{All}(E, (3, 4))$
\State $F,\space D,\space P \gets \textsc{NonZero}(A)$
\State $F_u,\space F_u^c \gets \textsc{UniqueWithCounts}(F)$
\State $C_u,\space C_u^i,\space C_u^c \gets \textsc{UniqueWithIndicesAndCounts}(F_u^c)$
\State $s \gets \textsc{Sum}(C_u^c)$
\If {$s \neq l$}
\State \textbf{raise} $\textsc{TimestampDetectionCountError}()$
\EndIf
\For {$ i = 1 $ \text{to} $\textsc{Len}(C_u^i) - 1$}
\If {$C_u^i[i] < C_u^i[i - 1]$}
\State \textbf{raise} $\textsc{NonDecreasingTimestampLenError}()$
\EndIf
\EndFor
\State $D \gets \textsc{AsType}(D, \textsc{UnicodeType})$
\State $S \gets \textsc{NDArray}(l, \textsc{IntegerType})$
\State $i_r \gets 0$
\For {$ i = 0 $ \text{to} $ \textsc{Len}(C_u) - 1$}
\State $c_u \gets C_u[i]$ \Comment{length of each timestamp in batch {$j$}}
\State $c_u^c \gets C_u^c[i]$ \Comment{number of {$c_u$}-length timestamps}
\State $n_l^r \gets c_u \times c_u^c$ \Comment{total digits spanning {$c_u^c$} timestamps}
\State $i_l \gets i_r$ \Comment{left index into {$c_u$}-length timestamps in {$D$}}
\State $i_r \gets i_l + n_l^r$ \Comment{right index into timestamps in {$D$}}
\State $P_l^r \gets P[i_l:i_r]$ \Comment{timestamp-grouped digit positions}
\State $P_l^r \gets \textsc{Reshape}(P_l^r, (c_u^c, c_u))$ \Comment{timestamp-wise {$P_l^r$}}
\State $P_l^r \gets \textsc{ArgSort}(P_l^r)$ \Comment{order positions increasingly}
\State $O \gets \textsc{Arange}(0, n_l^r, c_u)$ \Comment{define index offsets...}
\State $O \gets \textsc{ExpandDims}(O, 1)$ \Comment{...into {$D$} for batch {$j$}}
\State $P_l^r \gets \textsc{Add}(O, P_l^r)$ \Comment{shift indices {$P_l^r$} by offsets {$O$}}
\State $D_l^r \gets D_l^r[i_l:i_r][P_l^r]$ \Comment{ordered {$c_u$}-length timestamps}
\State $c_u^i \gets C_u^i[i]$ \Comment{first {$c_u$}-length timestamp index into {$D_l^r$}}
\For {$ j = c_u^i $ \text{to} $ c_u^i + c_u^c$} \Comment{concatenate {$c_u$} digits...}
\State $S[j] \gets \textsc{Join}(D_l^r[j - c_u^i])$ \Comment{...into one string}
\EndFor
\EndFor\\
\Return {$S$}
\end{algorithmic}
\end{algorithm} 

\subsection{Video Frame Timestamp Extraction}
The SHRP 2 videos are one type of what is referred to as supplemental data. Supplemental data are stored on file systems and not directly integrated into the RID. To integrate information derived from these raw data into the RID, a process named “conflation” with a geospatial database named the Linear Referencing System (LRS) has been defined. To conflate work zone scene detections, allowing them to be localized using existing GPS information for a given trip, the beginning and end timestamps of detected scenes needed to be identified. Because the RID did not already contain a mapping from frame numbers to timestamps for every video, the only way to temporally localize work zone scenes in videos was to directly extract the digital timestamps graphically overlaid on video frames.

A pseudocode for the algorithm that performs this extraction is outlined in Algorithm \ref{alg:convtc}:
Lines 1 and 2 define the number of timestamps, {$l$}, and the maximum number of digits per timestamp, {$n$}, in the input array of timestamps, {$T$}. {$n$} is given by the ratio of the maximum timestamp width, {$w$}, and timestamp height, {$h$}, because each timestamp digit image is \(16 \times 16\) pixels square. Lines 3 through 6 define {$M$} to be an array of binary image masks representing the ten Arabic numerals, and then prepare {$M$} to be compared for equality against each digit of each timestamp. The three color channels in {$T$}'s third dimension are collapsed into one grayscale channel and then converted to binary black and white to match the format of {$M$} in line 7. In lines 8 and 9, {$T$} is reshaped to match the dimensions of {$M$}. Lines 10 and 11 test each digit in each timestamp for equality with all ten Arabic numeral masks and produce a 3D array of shape \(l \times 10  \times n\), where one truth value exists for each numeral. Ideally, exactly one of the values will be True and the other nine False. Line 12 extracts matches using three 1D arrays to represent them, {$F$}, {$D$} and {$P$}. Each array contains indices into one of the dimensions of the array output by line 11. Conveniently, the three arrays have a semantic interpretation: the values in {$F$} represent the frame number from which each timestamp was extracted, {$D$} contains the numerical values of timestamp digits, and the values in {$P$} represent the position at which those digits occurred in their timestamp. From these three arrays of integers, we can construct string representations of what were previously images. 

\begin{algorithm}
\caption{BinarizeTimestampImages(\textit{T})}\label{alg:bints}
\begin{algorithmic}[1]
\State $T \gets \textsc{Average}(T, 2)$ \Comment{convert image to grayscale}
\State $t \gets 128$ \Comment{define binarization threshold}
\State $w \gets [255]$ \Comment{define white pixel value}
\State $b \gets [0]$ \Comment{define black pixel value}
\State $T \gets \textsc{Where}(T >= t, w, b)$ \Comment{binarize image}\\
\Return {$T$}
\end{algorithmic}
\end{algorithm}

At this point, our ability to exploit Numpy's broadcasting feature becomes constrained due to potential variability in the number of digits that represent a timestamp within a single video (consider for example the transition from \(t = 999965\) to \(t = 1000031\) for \(t \in T\)). Continuing in the interest of minimizing computation, lines 13 and 14 prepare the algorithm to iterate over batches of timestamps sharing a common number of digits rather than over each timestamp one at a time. This is achieved by ordering the values in F to be monotonically non-decreasing, resulting in {$C_u$}, identifying the index of the first occurrence of each numeral in the arrangement, {$C_u^i$}, and separately the number of occurrences of each numeral in {$C_u^c$}.

Before proceeding, two quality control checks are performed between lines 15 and 20 to determine if any one of the timestamps could not be read due to some form of distortion in the corresponding video frame. The first looks for evidence of any missing timestamps. The second checks for evidence of an existing timestamp with one or more of its digits missing. If either check is satisfied, a 5\% slower variant of the conversion algorithm that operates on each timestamp individually is run to identify the culprits and quality control them by synthesizing an artificial replacement timestamp. Because the resolution of link-level information in the LRS is much lower than the frame rate of videos, the error introduced by using synthetic timestamps is inconsequential. We do not present this alternative algorithm here. 

With the extracted timestamps validated, conversion to a string representation can proceed. Line 21 converts the individual digits from integers into string, and line 22 initializes a new n-dimensional array of length {$l$} in which the output strings will be placed. The variable {$i_r$}, which is used inside the loop starting on line 24 together with {$i_l$} to extract batches of the digits of equal-length timestamps, is initialized outside of the loop on line 23. For a detailed treatment of the string construction loop, please see the figure for Algorithm \ref{alg:convtc} starting at line 25. 

Unfortunately, the operation that concatenates individual digit strings of length one into a single timestamp string of length {$c_u$} does not have a broadcast implementation in Numpy. The library simply calls the Python equivalent function. Thus, this {$O$}({$l$}) operation could not be optimized.

\begin{figure*}[t]  
\begin{center}  
\begin{tabular}{cc}  
\includegraphics[width=7in]{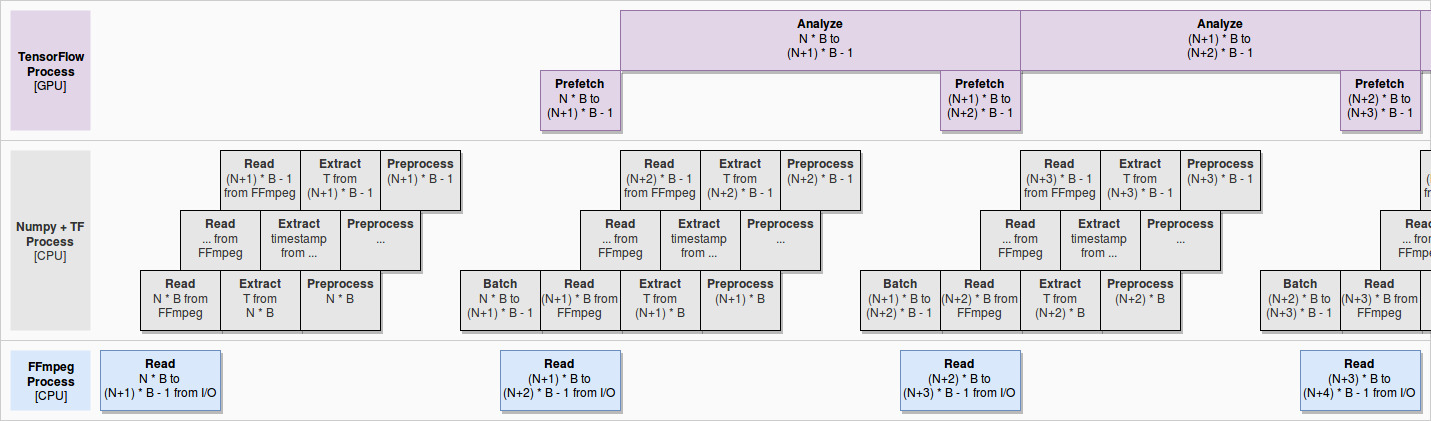}
\end{tabular}  
\caption{\label{fig:video-processing-pipeline}Video Processing Pipeline: An abstract diagram describing the concurrent nature of the pipeline through which video data is ingested and processed. In the SNVA development environment, we achieved total GPU saturation while processing batches of frames within a given video, and only a 2-3 second delay between the last batch of one video and the first batch of the next.}  
\end{center}  
\end{figure*}

\subsection{Input Pipeline Optimization}
Particular attention was given to the design of SNVA's video processing pipeline,  the core of the application. Much of the design was inspired by key features of the TF data ingestion API and by guidance in the TF documentation. Numpy and TF preprocessing are performed on the CPU in order to dedicate GPUs to the task of inference and maximize concurrency in the pipeline. An abstract graphical representation of the pipeline is presented in Figure \ref{fig:video-processing-pipeline}.

\subsubsection{Generator Functions for Continuous Video Frame Streaming}
The option of feeding the TF data ingestion pipeline using a generator function allowed us to pipe FFmpeg's output directly into TF, effectively handing the responsibility of optimizing the reading of frames from the pipe to TF. We as application developers could take advantage of TF's fast multi-threaded readers with minimal development effort. The function reads one frame from the FFmpeg buffer at a time, extracts the frame's timestamp and stores it in an array to be returned together with probabilities at the end of processing, then crops away extraneous edge pixels to maximize the sizes of image features after rescaling to the CNN's fixed input image size, and finally yields the cropped image to TF. The TF pipeline then applies model-specific preprocessing transformations to multiple frames concurrently, batches a specified number of frames together to be processed concurrently on the GPU, and lastly pre-loads the batch into GPU memory where it waits to start being processed immediately after the preceding batch finishes.

\subsubsection{Model- and Hardware-specific Video Frame Batching}
Determining the optimal batch size for a given hardware configuration is a task left to the SNVA's users. If too low, the GPU risks being under-utilized. If too high performance may suffer, perhaps because there is insufficient available memory to pre-fetch the entire batch of images. For example, we found in our experiments with MobilenetV2 that a batch size of 64 was optimal in terms of frames processed per second even though our development machine's GPU easily supported a batch size of 128. 

\subsubsection{Using TensorRT and the Channels-first Data Format}
Following TF and NVIDIA guidelines, we converted SNVA's models from the channels-last (NHWC) to the channels-first (NCHW) format in order to maximize inference performance. While we could not find and explicit explanation for the benefit, the use of NCHW is promoted in at least three locations in each of TF's and NVIDIA's documentation. Further, because SNVA takes advantage of TensorRT, and TensorRT only supported the NCHW format at the time of this writing, there was necessarily a transpose operation performed "under the hood" on each batch of images during processing. To avoid performing this computation on the GPU, and to gain the aforementioned unexplained benefits, we converted SNVA's models from NHWC to NCHW and added a transpose operation to the preprocessing pipeline that runs on CPU.  We observed a 1\% speed increase following the change, which added to the 10\% speed increase gained by utilizing TensorRT in the first place.

\subsubsection{Freezing TF Inference Graphs}
Here, again, we follow guidance from the TF documentation and convert the model files output in TF checkpoint format during training, in which contain large TF variable objects, to protobuf files in which all variables have been converted into constant values. As expected and intended, the performance gains are non-trivial.


\subsection{Software Development Environment}

SNVA was developed and alpha tested using an Alienware Area 51 workstation with a 10-core 3.00GHz Intel® Core™ i7-6950X CPU with hyperthreading, 64GB of DDR4 SDRAM, and two NVIDIA GeForce GTX 1080 Ti graphics cards with 11GB of GDDR5 RAM each. The machine ran the Ubuntu 16.04 operating system, with NVIDIA driver 396.24, CUDA 9.0, cuDNN 7.0, TensorRT 3.0.4 , TensorFlow 1.8.0, Docker 18.03.1-CE, and NVIDIA-Docker 2.0.3 installed. The Python and Numpy versions used were 3.5 and 1.14, respectively. 

When this setup was tested against all 31,535,862 video frames spanning 1,344 videos in the training pool, InceptionV3 inferred class labels at 826.24 fps on average over 10:40:04 hours, MobilenetV2 (with one video processor assigned to each GPU) averaged 1491.36 fps over 05:58:20 hours, and MobilenetV2 (with two video processors assigned to each GPU) averaged 1833.1 fps over 4:53:42 hours. RAM consumption on our development machine appeared to be safely bounded above by 3.75GB per active video analyzer.

\subsection{Production SNVA Environment}

The production environment to which SNVA was deployed included Dell PowerEdge C4130 servers featuring four NVIDIA Tesla V100 GPUs each. Given the use of Docker to deploy and run SNVA, the software environment in production is identical to that of development, with a marginal exception to the exact NVIDIA driver version. Ubuntu is also equal in version by construction. The videos are streamed from an NFS network share over a 16Gb/s link. The fastest SNVA configuration (four MobilenetV2 analyzers across two GPUs) consumed videos at an average of 3.5Gb/s when reading from an internal 7200RPM hard disk drive. While beta test results were not yet available at the time of this writing, these numbers gave us confidence that the V100s would also achieve full saturation.


\section{Future Work}\label{futurework}

\subsection{Precision-oriented Active Learning}

As was mentioned in section \ref{sec:datasetconstructionandmodeldevelopment}, work zone scenes represent a small percentage of the total data. It may therefore be feasible to include samples misclassified as  work zone scenes with high confidence in the labeling process once the model becomes sufficiently accurate. This means that in addition to model confidence on as yet unlabeled samples, and to the F0.5 score on the in-sample test set, precision on unlabeled samples can eventually become a candidate for measuring model performance. Recall that precision is the ratio of samples correctly classified as positive to all samples classified as positive. This would allow the unlabeled data to better indicate the model's potential efficacy on the target data set. Recall would be expected to improve though it would not be measured.

\subsection{Robust Multi-Frame Event Detection using Bidirectional Recurrent Neural Networks}

An early design of the SNVA application included a combination of a CNN as a work zone feature detector and a bidirectional RNN \cite{DBLP:journals/corr/abs-1303-5778} as a work zone event detector, where an event may span many frames and include intermediate frames that may not actually contain work zone features. Consider for example a research volunteer driving through a work zone in the center lane when there is heavy traffic, leading to periodic occlusions of equipment by other vehicles. While the SHRP 2 video data set, being static and historical rather than real-time and online, is an attractive candidate for the application of an RNN that incorporates information from preceding and following frames into its prediction for the current frame, we ruled out the use of such an architecture for two reasons. First, we anticipated that budgetary and time constraints would not support the level of effort required to perform annotation at the video level. We turned out to be right. Second, after observing the number of frames that had to be excluded from consideration when developing the training set due to anomalies and ambiguities, it was not clear how video-level annotation should be approached. As a research exercise not necessarily tied to SNVA, we intend to explore the feasibility to applying an RNN to SHRP 2 video.

\subsection{Other High-priority Target Scene Features}

As mentioned in the introduction, there exist a number of environmental conditions and roadway features that are not currently included in the RID but that researchers and other stakeholders have wanted to include in their studies dating back to the creation of the RID. Among them are traffic signal states and weather conditions. The SNVA application and the model development methodology presented in this paper can readily be extended to support detection of these features, and we intend to target them in future versions of SNVA.

\section{Conclusion}\label{conclusion}
In this paper we have introduced SNVA, the SHRP 2 NDS Video Analytics software application. SNVA adds to the Roadway Information Database the ability for transportation safety researchers to formulate and ask questions specific to construction zones transited by Naturalistic Driving Study volunteer participant drivers. While the application was still undergoing beta testing at the time of this writing, alpha testing in the development environment implied that RID query results would be accurate and exhaustive. We described the approaches followed in developing the application as well as the machine learning model used to infer scene classifications from images in detail. The motivations behind the project and potential benefits to both data science and transportation communities if successful were also discussed. Currently, SNVA targets only one type of information: work zone presence, but in the near future we intend to expand its capabilities to apply to weather events and traffic signal state. There are also opportunities to improve the application's robustness to fluctuations in the presence of features within an event region by feeding CNN feature vectors into a bidirectional recurrent neural network (likely an LSTM). The source code repository is publicly available at: \url{https://github.com/VolpeUSDOT/SNVA}  

\section*{Acknowledgment}
Mr. Abodo gratefully acknowledges David Kuehn, Charles Fay and Yusuf Mohamedshah of FHWA's Turner-Fairbank Highway Research Center (TFHRC),  Miguel Perez, Joel Anderson and Calvin Winkowski of VTTI, Thomas Karnowski of Oak Ridge National Laboratory (ORNL) and Omar Smadi of Iowa State University's Center for Transportation Research and Education (CTRE) for their guidance and support during the development of SNVA. He would also like to thank his research advisor Leonardo Bobadilla and fellow Motion, Robotics and Automation (MoRA) lab mates Md. Mahbubur Rahman, Tauhidul Alam and Sebasti\'{a}n Zanlongo for providing instructive opportunities to engage in computer science research activities as an undergraduate at Florida International University's School of Computing and Information Sciences. This work was funded under Inter-Agency Agreement HW53A100 between the Volpe Center and FHWA.

\bibliographystyle{IEEEtran}
\bibliography{bib/alt.bib}
\end{document}